\documentclass[conference,a4paper]{IEEEtran}
\IEEEoverridecommandlockouts
\usepackage{cite}
\usepackage{amsmath,amssymb,amsfonts}
\usepackage{algorithmic}
\usepackage{booktabs, makecell, multirow, tabularx}
\usepackage{graphicx}
\usepackage{textcomp}
\usepackage{xcolor}
\usepackage[hidelinks]{hyperref} 
\usepackage{siunitx}
\def\BibTeX{{\rm B\kern-.05em{\sc i\kern-.025em b}\kern-.08em
    T\kern-.1667em\lower.7ex\hbox{E}\kern-.125emX}}

\usepackage{multicol}
\usepackage{subcaption}

\begin{document}

\title{FERT: Real-Time Facial Expression Recognition with Short-Range FMCW Radar\\

}

 \author{
    \IEEEauthorblockN{Sabri Mustafa Kahya$^{\star}$ \qquad Muhammet Sami Yavuz$^{\star}$ \qquad Eckehard Steinbach$^{\star}$}
    \vspace{0.2cm}
    \IEEEauthorblockA{\IEEEauthorrefmark{1}Technical University of Munich, School of Computation, Information and Technology, Department of Computer Engineering, \\Chair of Media Technology,\\ 
    Munich Institute of Robotics and Machine Intelligence (MIRMI),\\
    Munich, Germany\\
    \{mustafa.kahya, sami.yavuz, eckehard.steinbach\}@tum.de}

} 
\maketitle

\begin{abstract}
This study proposes a novel approach for real-time facial expression recognition utilizing short-range Frequency-Modulated Continuous-Wave (FMCW) radar equipped with one transmit (Tx), and three receive (Rx) antennas. The system leverages four distinct modalities simultaneously: Range-Doppler images (RDIs), micro range-Doppler Images (micro-RDIs), range azimuth images (RAIs), and range elevation images (REIs). Our innovative architecture integrates feature extractor blocks, intermediate feature extractor blocks, and a ResNet block to accurately classify facial expressions into smile, anger, neutral, and no-face classes. Our model achieves an average classification accuracy of 98.91\% on the dataset collected using a 60 GHz short-range FMCW radar. The proposed solution operates in real-time in a person-independent manner, which shows the potential use of low-cost FMCW radars for effective facial expression recognition in various applications.
\end{abstract}

\begin{IEEEkeywords}
Facial expression recognition, short-range FMCW radar, deep-learning
\end{IEEEkeywords}

\section{Introduction}
Facial Expression Recognition (FER) is a crucial technique for identifying emotional states in humans by analyzing their facial expressions \cite{SurveyFER}. By accurately detecting and analyzing these expressions, FER finds practical applications in enhancing Human-Computer Interaction (HCI) systems\cite{UFace}, virtual reality experiences \cite{vrFER}, and digital entertainment \cite{IntroFER,digitalEntFER}. Despite the prevalence of vision-based approaches in FER \cite{visionFER,visionFER2,visionFER3}, they are hindered by factors such as lighting and occlusion. Additionally, they are not privacy-preserving.

Short-range radars have gained popularity in both academia and industry due to their robustness in various environmental conditions and their privacy advantages. They are primarily used for indoor applications, including human presence detection, human activity classification, people counting, gesture recognition, and out-of-distribution (OOD) detection \cite{human_presence7,kahya2023hood,human_activity3,kahya2023harood,gesture_recog,RB-OOD,MCROOD}. Additionally, they are employed for vital sign monitoring, such as heartbeat estimation \cite{heartbeat_est}, and for capturing facial features in authentication systems \cite{dnn-based,cnn-based,one-shot,point-cloud-face,kahya2024food}. They are also used in facial expression recognition systems. \cite{mmFER} introduces a mmWave radar-based FER system capable of capturing subtle facial muscle movements directly from raw mmWave signals, enabling recognition of seven standard facial expressions across multiple users. The system employs a novel dual-locating approach to accurately position subjects' faces in space and extract facial muscle movements from noisy raw signals. \cite{mm3DFace } employs 3D range angle spectrum (range-azimuth-elevation) of mmWave radar with three transmit (Tx) and four receive (Rx) antennas for 3D facial reconstruction, capturing changes in facial expressions using triplet loss embedding.

While FER offers a fundamental method for extracting emotional states from facial expressions, some radar-based studies utilizing vital signs for emotion recognition provide additional insights into the complex interplay of physiological responses. Continuous-wave (CW) Doppler radar, for instance, has been instrumental in classifying emotions by capturing respiratory signal features from video stimuli \cite{emot_recog1,emot_recog2}. Similarly, millimeter-wave radar can monitor heartbeats and/or respiration, extracting features that empower deep/machine learning models in recognizing emotions\cite{emot_recog3,emot_recog6,mmWaveEmot_Recog}. Additionally, a non-contact multimodal system leverages a CW Doppler radar and a camera to capture respiratory and heartbeat signals for enhanced emotion recognition\cite{emot_recog4}.  

In our work, we introduce FERT, a novel facial expression recognition system using short-range FMCW radar. This system is designed to accurately classify smile, anger, neutral, and no-face expressions in a person-independent and real-time manner. Unlike camera-based solutions, our radar-based solution aims to model the specific movements of the muscles on the face while performing the specific expression. Therefore, in its unique pipeline, FERT benefits from Range-Doppler Images (RDIs), micro-RDIs, Range Azimuth Images (RAIs), and Range Elevation Images (REIs) simultaneously. The system includes dedicated feature extractor blocks for each modality, ensuring comprehensive feature extraction from all four data types. After initial feature extraction, the intermediate features of similar modalities are combined: RDIs are merged with micro-RDIs, and RAIs with REIs. The combined intermediate features are fed into two separate intermediate feature extractors—one for RDIs \& micro-RDIs, and another for RAIs \& REIs. Subsequently, the outputs from these intermediate feature extractors are further combined to integrate all modalities at a certain point. The unified features are then fed into a ResNet block for the classification task. This sophisticated approach allows FERT to achieve high accuracy and real-time performance, making it a robust solution for facial expression recognition.

\section{Radar Configuration \& Pre-processing}

This study utilizes Infineon's BGT60TR13C $\SI{60}{\giga\hertz}$ FMCW radar chipset, which has one Tx antenna and three Rx antennas. The radar configuration is detailed in Table \ref{tab:radar_conf}. The Tx antenna emits \(N_c\) chirp signals, which are reflected and captured by the Rx antennas. These signals are mixed and low-pass filtered to derive the intermediate frequency (IF) signal. The IF signal is then digitized to produce the raw Analog-to-Digital Converter (ADC) data, organized as \(N_{Rx} \times N_c \times N_s\), where \(N_c\) represents slow time and \(N_s\) fast time samples.

We generate macro and micro-RDIs, RAIs, and REIs as input to our model. Macro RDI is produced by first applying \textbf{Range-FFT} to the fast-time signal for range extraction, followed by mean removal and moving target identification (\textbf{MTI}) to eliminate static targets. \textbf{Doppler-FFT} then processes the slow-time signal to capture phase variations, producing the macro RDI. Micro-RDI is generated in a similar manner using \textbf{Range-FFT} on range data. Noise reduction is achieved by stacking eight range spectrograms and removing means from both the fast and slow time signals. Enhanced target detection is facilitated through \textbf{Sinc} filtering, and a final \textbf{Doppler-FFT} along the slow-time dimension completes the micro-RDI generation. For RAI and REI generation, we use the micro-RDI and apply the Direction of Arrival (DoA) estimation with the \textbf{Capon} beamforming algorithm. RAI and REI are obtained using two antennas, given our chipset's L-shaped configuration in the Rx antennas. The two antennas on one side of the L are used for RAI, while the other two antennas on the other side of the L are used for REI. As a final preprocessing step, we applied \textbf{E-RESPD} \cite{kahya2023hood} to enhance the capture of facial movements on both RDIs, RAI and REI.

\begin{table}[h]
    \caption{\small FMCW Radar Configuration Parameters }
    \centering
     \footnotesize  
    \begin{tabular}{@ {\extracolsep{10pt}} ccc}
    \toprule

    \centering
    Configuration name & Symbol & Value \\
    \midrule
    Number of transmit antennas & $N\textsubscript{Tx}$  & 1  \\
    Number of receive antennas & $N\textsubscript{Rx}$  & 3  \\
    Chirps per frame & $N\textsubscript{c}$ & 64  \\
    Samples per chirp & $N\textsubscript{s}$  & 128 \\
    Frame period & T\textsubscript{f} & 50 \si{\ms}  \\
    Chirp to chirp time & $T\textsubscript{cc}$ & 391.55 \si{\us} \\
    Bandwidth & $B$ & $\SI{1}{\giga\hertz}$\\ 
    \bottomrule
    \end{tabular}

    \label{tab:radar_conf}
\end{table}
\section{Problem Statement and FERT}

\begin{figure*}[htbp]
\centerline{\includegraphics[width=1\linewidth]{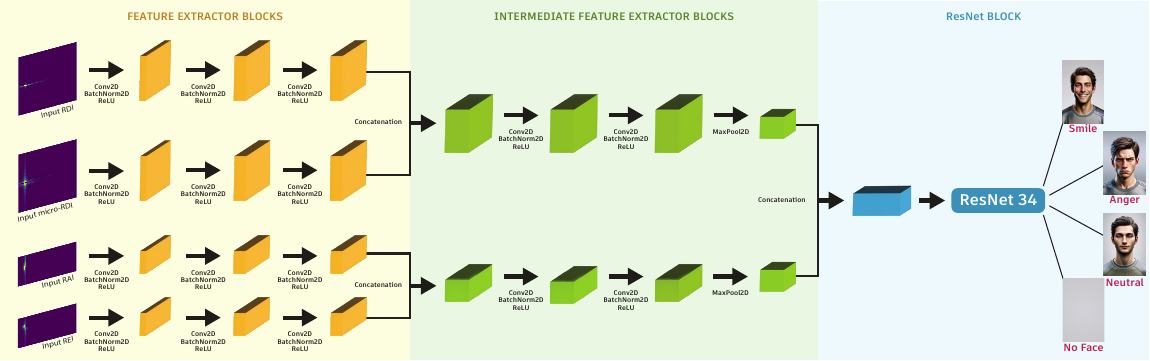}}
\caption{  This figure shows the overall architecture of FERT. The section with the yellow background consists of four feature extractor blocks for RDI, micro-RDI, RAI, and REI inputs, respectively. The section with the green background contains two intermediate feature extractor blocks for merged modalities: RDI \& micro-RDI and RAI \& REI. The section with the blue background is the ResNet block, which integrates all modalities for the final classification. All blocks are trained simultaneously to achieve highly accurate facial expression recognition.}
\label{fig:pipline}
\end{figure*}

Recognizing facial expressions is crucial for understanding a person's emotional state, which is essential for smart applications to deliver appropriate content. For instance, a driver's emotional state helps a smart car suggest the next action, while a user's emotional state influences the content provided by a smart glass. Therefore, FERT proposes a real-time, robust facial expression recognition pipeline using low-cost, easily integrable FMCW radar in a privacy-preserving manner. 

FERT consists of seven distinct blocks: four feature extractor blocks for RDI, micro-RDI, RAI, and REI modalities; two intermediate feature extractor blocks for paired modalities; and a ResNet34 \cite{resnet} block that takes input from all modalities.

The \textbf{feature extractors} share the same architecture, each consisting of three consecutive convolutional layers. Each layer includes 2D convolution, 2D batch normalization, and ReLU activation. The first layer uses 8 filters, while the second and third layers each use 16 filters, all with a $3 \times 3$  kernel size. These feature extractor blocks are trained only with their respective modalities to learn their low-level features. After these blocks, similar modalities are merged: RDIs with micro-RDIs, and RAIs with REIs.

The two \textbf{intermediate feature extractor} blocks handle the merged modalities, consisting of two consecutive convolutional layers. Each layer sequentially applies 2D convolution, 2D batch normalization, ReLU activation, and 2D max pooling, using 32 filters with a $3 \times 3$  kernel size. These blocks learn the intermediate features of the merged modalities. The outputs of these blocks are then merged to integrate all modalities.

The \textbf{ResNet} block takes this merged output, which includes information from all modalities, as input. It learns from all modalities and performs the final classification. For a comprehensive view of the entire pipeline, please refer to Figure \ref{fig:pipline}.

The entire network is trained simultaneously using cross-entropy loss and an SGD optimizer. Our architecture is a fast learner, so it is trained only three epochs to reach the final results.

\section{Experiments and Results}
Our experiments are conducted on a system equipped with an NVIDIA GeForce RTX 3070 GPU, an Intel Core i7-11800H CPU, and 32GB of DDR4 RAM.
\subsection{Dataset and Evaluation}

In this research, we collected a facial expression dataset using Infineon's BGT60TR13C $\SI{60}{\giga\hertz}$ FMCW radar chipset over one month. The radar sensor was positioned 25 cm away to capture the individual's facial expressions, with participants placing their chins on a table for alignment. Each recording session lasted between 1 to 5 minutes. We captured three primary facial expressions—anger, smile, and neutral—and one no-face class. The anger expression involved the person clenching their teeth, furrowing their brow, and narrowing their eyes. The smile was captured by having the person smile broadly, showing their teeth, and raising their cheeks. The neutral expression was maintained without any specific facial movement. We repeated the same movements for the corresponding facial expressions during each recording session. Importantly, individuals did not wear any accessories on their faces, such as glasses, during the recordings. The data was collected in various rooms to introduce environmental variability. Our dataset consists of four classes: anger, smile, neutral facial expressions, and no face in front of the radar. The training dataset includes 240330 data frames, while the testing dataset contains 105569 data frames, both sourced from two individuals in a balanced way for each expression. For real-time experiments, we also tested on individuals who did not exist in either training or test sets. Written consent has been obtained from all participants involved in the research.

\begin{figure}[ht]

    \centering

    \includegraphics[width=0.8\columnwidth]{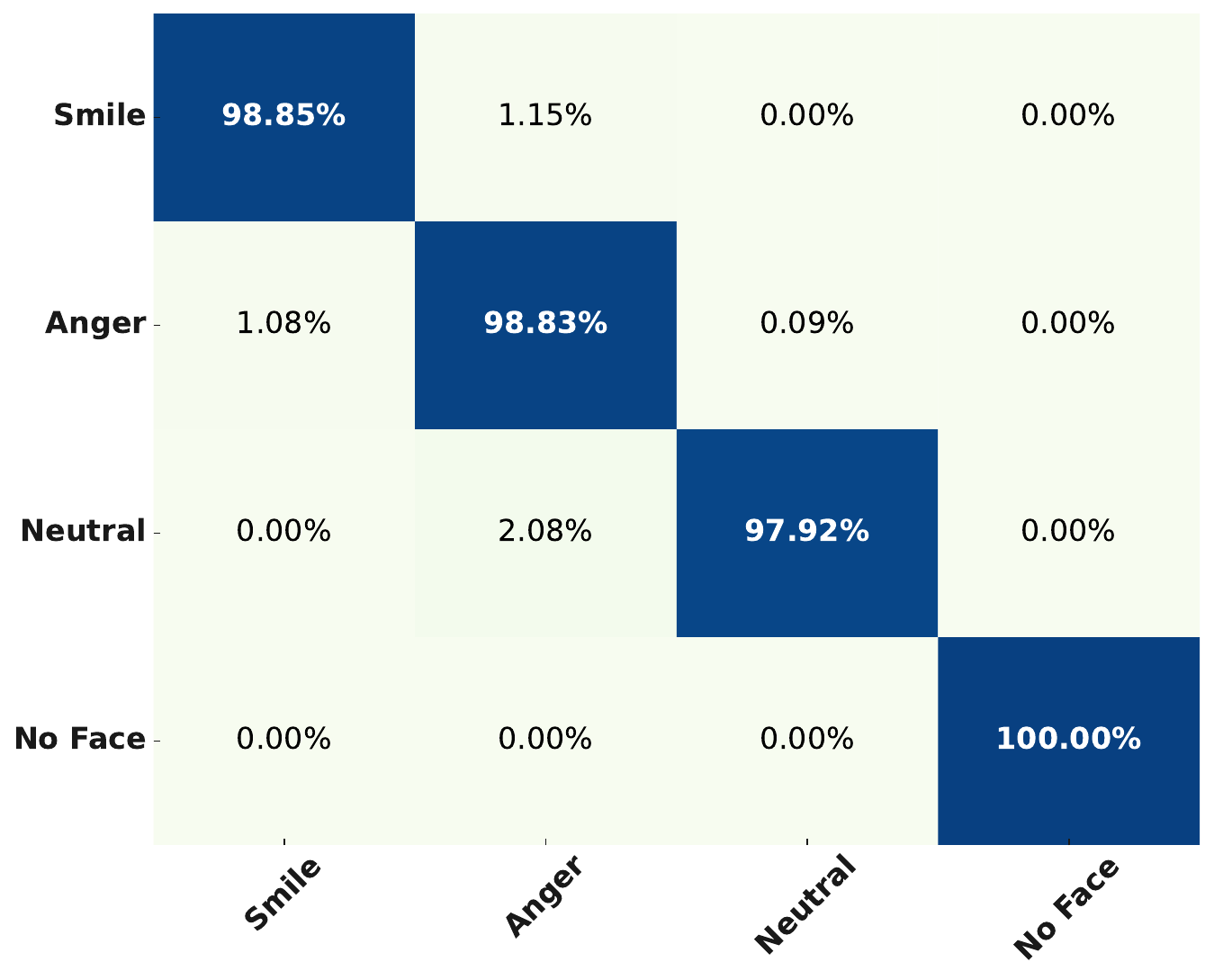}
    
     \caption{\small Confusion matrix to demonstrate the classification performance of FERT.}
     \label{fig:conf_matrix_FERT}
\end{figure}

In our experiments, we always use completely different recordings for training and testing. To avoid overfitting to specific locations where the radar is positioned, we use distinct locations for experimentation. Our solution achieves highly accurate classification results of 98.85\%, 98.83\%, 97.92\%, and 100.0\% for smile, anger, neutral, and no-face classes, respectively. To demonstrate the network's generalization capability, we also test our pipeline in real-time with participants not involved in the training set. This ensures that FERT performs accurately regardless of location or participant variability. Additionally, we provide a confusion matrix in Figure \ref{fig:conf_matrix_FERT} for facial expression classification to present a detailed analysis. Our real-time demo videos are available here\footnote{\href{https://youtu.be/N2z488zxauQ}{https://youtu.be/N2z488zxauQ}}.

\subsection{Ablation}
In our pre-processing step, inspired by \cite{kahya2023hood}, we use E-RESPD approach. E-RESPD is a simple yet effective approach that uses consecutive 200 frames simultaneously instead of performing framewise usage. The idea also positively affects the overall classification performance of our pipeline. Therefore, we perform an ablation study to demonstrate the impact of the E-RESPD approach. Please see Table \ref{classification-res}.

\begin{table}[ht]

\small
\caption{\small Ablation study for E-RESPD. }
\centering
\begin{tabular}{@{\extracolsep{\fill}}ccccc}
\toprule   
{} &{}&\multicolumn{2}{c}{Accuracy } \\

 \cmidrule{2-5} 
\centering 
   & SMILE  &ANGER & NEUTRAL & NO FACE\\ 
\midrule
FERT w/o E-RESPD &80.97 &87.75 & 75.65& 99.97\\
 FERT   & \textbf{98.85}& \textbf{98.83}&\textbf{97.92} & \textbf{100.0}   \\

\bottomrule
\end{tabular}
\label{classification-res}
\end{table}

\section{Conclusion}
In conclusion, this study presents a novel real-time facial expression recognition method using a low-resolution FMCW radar system. By employing RDI, micro-RDI, RAI, and REI modalities, our approach effectively captures the necessary features for distinguishing between different facial expressions. Our architecture's feature extractor, intermediate feature extractor, and ResNet blocks enable robust classification performance, achieving a 98.91\% average accuracy on the collected dataset. Our system's real-time capability and high accuracy underscore the viability of using low-cost and privacy-preserving radar technology for facial expression recognition in areas such as human-computer interaction, virtual reality, and digital entertainment.

\bibliographystyle{ieeetr}

\clearpage

\begin{thebibliography}{10}

\bibitem{SurveyFER}
Shan Li and Weihong Deng,
\newblock ``Deep facial expression recognition: A survey,''
\newblock {\em IEEE Transactions on Affective Computing}, vol. 13, no. 3, pp. 1195--1215, 2022.

\bibitem{UFace}
Shuning Wang, Linghui Zhong, Yongjian Fu, Lili Chen, Ju~Ren, and Yaoxue Zhang,
\newblock ``Uface: Your smartphone can" hear" your facial expression!,''
\newblock {\em Proceedings of the ACM on Interactive, Mobile, Wearable and Ubiquitous Technologies}, vol. 8, no. 1, pp. 1--27, 2024.

\bibitem{vrFER}
Bita Houshmand and Naimul Mefraz~Khan,
\newblock ``Facial expression recognition under partial occlusion from virtual reality headsets based on transfer learning,''
\newblock in {\em 2020 IEEE Sixth International Conference on Multimedia Big Data (BigMM)}, 2020, pp. 70--75.

\bibitem{IntroFER}
Feifei Zhang, Tianzhu Zhang, Qirong Mao, and Changsheng Xu,
\newblock ``Joint pose and expression modeling for facial expression recognition,''
\newblock in {\em 2018 IEEE/CVF Conference on Computer Vision and Pattern Recognition}, 2018, pp. 3359--3368.

\bibitem{digitalEntFER}
Suman Saha, Rajitha Navarathna, Leonhard Helminger, and Romann~M Weber,
\newblock ``Unsupervised deep representations for learning audience facial behaviors,''
\newblock in {\em Proceedings of the IEEE Conference on Computer Vision and Pattern Recognition Workshops}, 2018, pp. 1132--1137.

\bibitem{visionFER}
Fanglei Xue, Qiangchang Wang, Zichang Tan, Zhongsong Ma, and Guodong Guo,
\newblock ``Vision transformer with attentive pooling for robust facial expression recognition,''
\newblock {\em IEEE Transactions on Affective Computing}, 2022.

\bibitem{visionFER2}
Yahui Nan, Jianguo Ju, Qingyi Hua, Haoming Zhang, and Bo~Wang,
\newblock ``A-mobilenet: An approach of facial expression recognition,''
\newblock {\em Alexandria Engineering Journal}, vol. 61, no. 6, pp. 4435--4444, 2022.

\bibitem{visionFER3}
Andrey~V. Savchenko, Lyudmila~V. Savchenko, and Ilya Makarov,
\newblock ``Classifying emotions and engagement in online learning based on a single facial expression recognition neural network,''
\newblock {\em IEEE Transactions on Affective Computing}, vol. 13, no. 4, pp. 2132--2143, 2022.

\bibitem{human_presence7}
Prateek Nallabolu, Li~Zhang, Hong Hong, and Changzhi Li,
\newblock ``Human presence sensing and gesture recognition for smart home applications with moving and stationary clutter suppression using a 60-ghz digital beamforming fmcw radar,''
\newblock {\em IEEE Access}, vol. 9, pp. 72857--72866, 2021.

\bibitem{kahya2023hood}
Sabri~Mustafa Kahya, Muhammet~Sami Yavuz, and Eckehard Steinbach,
\newblock ``Hood: Real-time robust human presence and out-of-distribution detection with low-cost fmcw radar,''
\newblock {\em arXiv}, 2023.

\bibitem{human_activity3}
Thomas Stadelmayer, Markus Stadelmayer, Avik Santra, Robert Weigel, and Fabian Lurz,
\newblock ``Human activity classification using mm-wave fmcw radar by improved representation learning,''
\newblock in {\em Proceedings of the 4th ACM Workshop on Millimeter-Wave Networks and Sensing Systems}, New York, NY, USA, 2020, mmNets'20, Association for Computing Machinery.

\bibitem{kahya2023harood}
Sabri~Mustafa Kahya, Muhammet Sami~Yavuz, and Eckehard Steinbach,
\newblock ``Harood: Human activity classification and out-of-distribution detection with short-range fmcw radar,''
\newblock in {\em ICASSP 2024 - 2024 IEEE International Conference on Acoustics, Speech and Signal Processing (ICASSP)}, 2024, pp. 6950--6954.

\bibitem{gesture_recog}
Souvik Hazra and Avik Santra,
\newblock ``Robust gesture recognition using millimetric-wave radar system,''
\newblock {\em IEEE Sensors Letters}, vol. 2, no. 4, pp. 1--4, 2018.

\bibitem{RB-OOD}
Sabri~Mustafa Kahya, Muhammet~Sami Yavuz, and Eckehard Steinbach,
\newblock ``Reconstruction-based out-of-distribution detection for short-range fmcw radar,''
\newblock in {\em 2023 31st European Signal Processing Conference (EUSIPCO)}, 2023, pp. 1350--1354.

\bibitem{MCROOD}
Sabri~Mustafa Kahya, Muhammet Sami~Yavuz, and Eckehard Steinbach,
\newblock ``Mcrood: Multi-class radar out-of-distribution detection,''
\newblock in {\em ICASSP 2023 - 2023 IEEE International Conference on Acoustics, Speech and Signal Processing (ICASSP)}, 2023, pp. 1--5.

\bibitem{heartbeat_est}
Muhammad Arsalan, Avik Santra, and Christoph Will,
\newblock ``Improved contactless heartbeat estimation in fmcw radar via kalman filter tracking,''
\newblock {\em IEEE Sensors Letters}, vol. 4, no. 5, pp. 1--4, 2020.

\bibitem{dnn-based}
Hae-Seung Lim, Jaehoon Jung, Jae-Eun Lee, Hyung-Min Park, and Seongwook Lee,
\newblock ``Dnn-based human face classification using 61 ghz fmcw radar sensor,''
\newblock {\em IEEE Sensors Journal}, vol. 20, no. 20, pp. 12217--12224, 2020.

\bibitem{cnn-based}
J.~Kim, J.‐E Lee, H.‐S Lim, and S.~Lee,
\newblock ``Face identification using millimetre-wave radar sensor data,''
\newblock {\em Electronics Letters}, vol. 56, 08 2020.

\bibitem{one-shot}
Ha-Anh Pho, Seongwook Lee, Vo-Nguyen Tuyet-Doan, and Yong-Hwa Kim,
\newblock ``Radar-based face recognition: One-shot learning approach,''
\newblock {\em IEEE Sensors Journal}, vol. 21, no. 5, pp. 6335--6341, 2021.

\bibitem{point-cloud-face}
Youxuan Zhong, Chun Yuan, Yi~Zou, and Heng Yao,
\newblock ``Face recognition based on point cloud data captured by low-cost mmwave radar sensors,''
\newblock in {\em 2023 IEEE 13th Annual Computing and Communication Workshop and Conference (CCWC)}, 2023, pp. 0074--0083.

\bibitem{kahya2024food}
Sabri~Mustafa Kahya, Boran~Hamdi Sivrikaya, Muhammet~Sami Yavuz, and Eckehard Steinbach,
\newblock ``Food: Facial authentication and out-of-distribution detection with short-range fmcw radar,''
\newblock {\em arXiv}, 2024.

\bibitem{mmFER}
Xi~Zhang, Yu~Zhang, Zhenguo Shi, and Tao Gu,
\newblock {\em mmFER: Millimetre-wave Radar based Facial Expression Recognition for Multimedia IoT Applications},
\newblock Association for Computing Machinery, New York, NY, USA, 2023.

\bibitem{mm3DFace}
Jiahong Xie, Hao Kong, Jiadi Yu, Yingying Chen, Linghe Kong, Yanmin Zhu, and Feilong Tang,
\newblock ``mm3dface: Nonintrusive 3d facial reconstruction leveraging mmwave signals,''
\newblock in {\em Proceedings of the 21st Annual International Conference on Mobile Systems, Applications and Services}, 2023, pp. 462--474.

\bibitem{emot_recog1}
Qian Gao, Li~Zhang, Jiaming Yan, Heng Zhao, Chuanwei Ding, Hong Hong, and Xiaohua Zhu,
\newblock ``Non-contact emotion recognition via cw doppler radar,''
\newblock in {\em 2018 Asia-Pacific Microwave Conference (APMC)}, 2018, pp. 1468--1470.

\bibitem{emot_recog2}
Carolina Gouveia, Ana Tomé, Filipa Barros, Sandra~C. Soares, José Vieira, and Pedro Pinho,
\newblock ``Study on the usage feasibility of continuous-wave radar for emotion recognition,''
\newblock {\em Biomedical Signal Processing and Control}, vol. 58, pp. 101835, 2020.

\bibitem{emot_recog3}
Xiaochao Dang, Zetong Chen, and Zhanjun Hao,
\newblock ``Emotion recognition method using millimetre wave radar based on deep learning,''
\newblock {\em IET Radar, Sonar \& Navigation}, vol. 16, no. 11, pp. 1796--1808, 2022.

\bibitem{emot_recog6}
Huanpu Yin, Shuhui Yu, Yingshuo Zhang, Anfu Zhou, Xin Wang, Liang Liu, Huadong Ma, Jianhua Liu, and Ning Yang,
\newblock ``Let iot know you better: User identification and emotion recognition through millimeter-wave sensing,''
\newblock {\em IEEE Internet of Things Journal}, vol. 10, no. 2, pp. 1149--1161, 2023.

\bibitem{mmWaveEmot_Recog}
Kelong Zeng and Guangyuan Liu,
\newblock ``Emotion recognition based on millimeter wave radar,''
\newblock in {\em Proceedings of the 2023 3rd International Conference on Bioinformatics and Intelligent Computing}, 2023, pp. 232--236.

\bibitem{emot_recog4}
Li~Zhang, Chang-Hong Fu, Hong Hong, Biao Xue, Xuemei Gu, Xiaohua Zhu, and Changzhi Li,
\newblock ``Non-contact dual-modality emotion recognition system by cw radar and rgb camera,''
\newblock {\em IEEE Sensors Journal}, vol. 21, no. 20, pp. 23198--23212, 2021.

\bibitem{resnet}
Kaiming He, Xiangyu Zhang, Shaoqing Ren, and Jian Sun,
\newblock ``Deep residual learning for image recognition,''
\newblock in {\em IEEE Conference on Computer Vision and Pattern Recognition (CVPR)}, 2016.

\end{thebibliography}

\end{document}